\documentclass[lettersize,journal,twoside]{IEEEtran}
\usepackage{amsmath,amsfonts,amssymb}
\usepackage{algpseudocode}
\usepackage{algorithm}
\usepackage{xcolor}
\usepackage{algpseudocode} %
\usepackage{array}
\usepackage[caption=false,font=normalsize,labelfont=sf,textfont=sf]{subfig}
\usepackage{textcomp}
\usepackage{stfloats}
\usepackage{url}
\usepackage{verbatim}
\usepackage{graphicx}
\usepackage{cite}
\usepackage{enumitem}
\usepackage{booktabs}
\usepackage{graphicx}
\usepackage{caption}

\begin{document}

\title{\spaceskip=0.3em plus 0.1em minus 0.1em IG-RFT: An Interaction-Guided RL Framework for VLA Models in Long-Horizon Robotic Manipulation}

\author{Zhian Su, Weijie Kong, Haonan Dong, and Huixu Dong,~\IEEEmembership{Member,~IEEE}
\thanks{Zhian Su, Weijie Kong, Haonan Dong, and Huixu Dong are with the Grasp Laboratory, Mechanical Engineering Department, Zhejiang University, Hangzhou 310027, China. (Corresponding author: Huixu Dong; e-mail: huixudong@zju.edu.cn).}
\thanks{All authors are also with Torch Kernel Co., Ltd., Hangzhou 310000, China.}
\thanks{This work has been submitted to the IEEE for possible publication. Copyright may be transferred without notice, after which this version may no longer be accessible.}
\vspace{-2mm}
}

\markboth{arXiv preprint}{Su et al.: IG-RFT: An Interaction-Guided RL Framework for VLA Models in Long-Horizon Robotic Manipulation}

\maketitle


\begin{abstract}
Vision-Language-Action (VLA) models have demonstrated significant potential for generalist robotic policies; however, they struggle to generalize to long-horizon complex tasks in novel real-world domains due to distribution shifts and the scarcity of high-quality demonstrations. Although reinforcement learning (RL) offers a promising avenue for policy improvement, applying it to real-world VLA fine-tuning faces challenges regarding exploration efficiency, training stability, and sample cost. To address these issues, we propose \textbf{IG-RFT}, a novel \textbf{I}nteraction-\textbf{G}uided \textbf{R}einforced \textbf{F}ine-\textbf{T}uning system designed for flow-based VLA models. Firstly, to facilitate effective policy optimization, we introduce Interaction-Guided Advantage Weighted Regression (IG-AWR), an RL algorithm that dynamically modulates exploration intensity based on the robot's interaction status. Furthermore, to address the limitations of sparse or task-specific rewards, we design a novel hybrid dense reward function that integrates the trajectory-level reward and the subtask-level reward. \textbf{Finally}, we construct a three-stage RL system comprising SFT, Offline RL, and Human-in-the-Loop RL for fine-tuning VLA models. Extensive real-world experiments on four challenging long-horizon tasks demonstrate that IG-RFT achieves an average success rate of \textbf{85.0\%}, significantly outperforming SFT (18.8\%) and standard Offline RL baselines (40.0\%). Ablation studies confirm the critical contributions of IG-AWR and hybrid reward shaping. In summary, our work establishes and validates a novel reinforced fine-tuning system for VLA models in real-world robotic manipulation.
\end{abstract}


\section{Introduction}
\IEEEPARstart{R}{obotic} manipulation for specific real-world applications has become increasingly mature\cite{dong2025enabling,su2025construction,lin2025coarse}. In comparison, generalist robotic policies are still not well-developed\cite{fei2025libero}. Recently, Vision-Language-Action (VLA) models, as a prominent paradigm for generalist policies\cite{black2410pi0, gr00tn1_2025, black2025pi_, shukor2025smolvla, zhai2025igniting}, have shown impressive capabilities and potential, owing to large-scale pre-training via behavior cloning. Existing VLA models have limited generalization, failing to achieve zero-shot transfer to novel real-world domains\cite{yakefu2025robochallenge}. Consequently, fine-tuning VLA models using domain-specific demonstrations is an indispensable procedure\cite{kim2025fine, zheng2025x}. The predominant domain adaptation approach for VLA models is supervised fine-tuning (SFT), which is incapable of effectively evaluating and utilizing demonstrations, thereby imposing stringent requirements on both the quality and quantity of training data. However, during data collection, sub-optimal data is frequently generated due to variations in operator proficiency, system errors, and discontinuous actions.

In contrast, reinforcement learning (RL) can explore, evaluate and utilize data more effectively to improve policies, exhibiting substantial potential for VLA domain adaptation\cite{chen2025pi_,yu2025rlinf,ye2025vla,xiao2025self,deng2025survey}. Unlike multi-environment parallel RL deployed in simulation\cite{wang2025bin,wang2025sa}, VLA models require direct physical interaction in real-world robotic tasks. This requires high sample efficiency and risk-aware exploration, rendering the direct implementation of on-policy RL infeasible. In real-world scenarios, deploying off-policy RL via Human-in-the-Loop can mitigate compounding errors caused by distribution shift, thereby guiding policy learning by timely corrections, yielding remarkable performance\cite{luo2025precise,chen2025conrft,physical2025pi}. However, these existing real-world RL systems use sparse rewards, which mainly causes unstable training in the online stage and leads to credit assignment problems. Furthermore, in long-horizon complex tasks, the variance of state distributions differs significantly across sub-stages; applying uniform exploration noise throughout the entire cycle results in inefficient resource utilization.

\begin{figure*}[t]
    \centering
    \includegraphics[width=2\columnwidth]{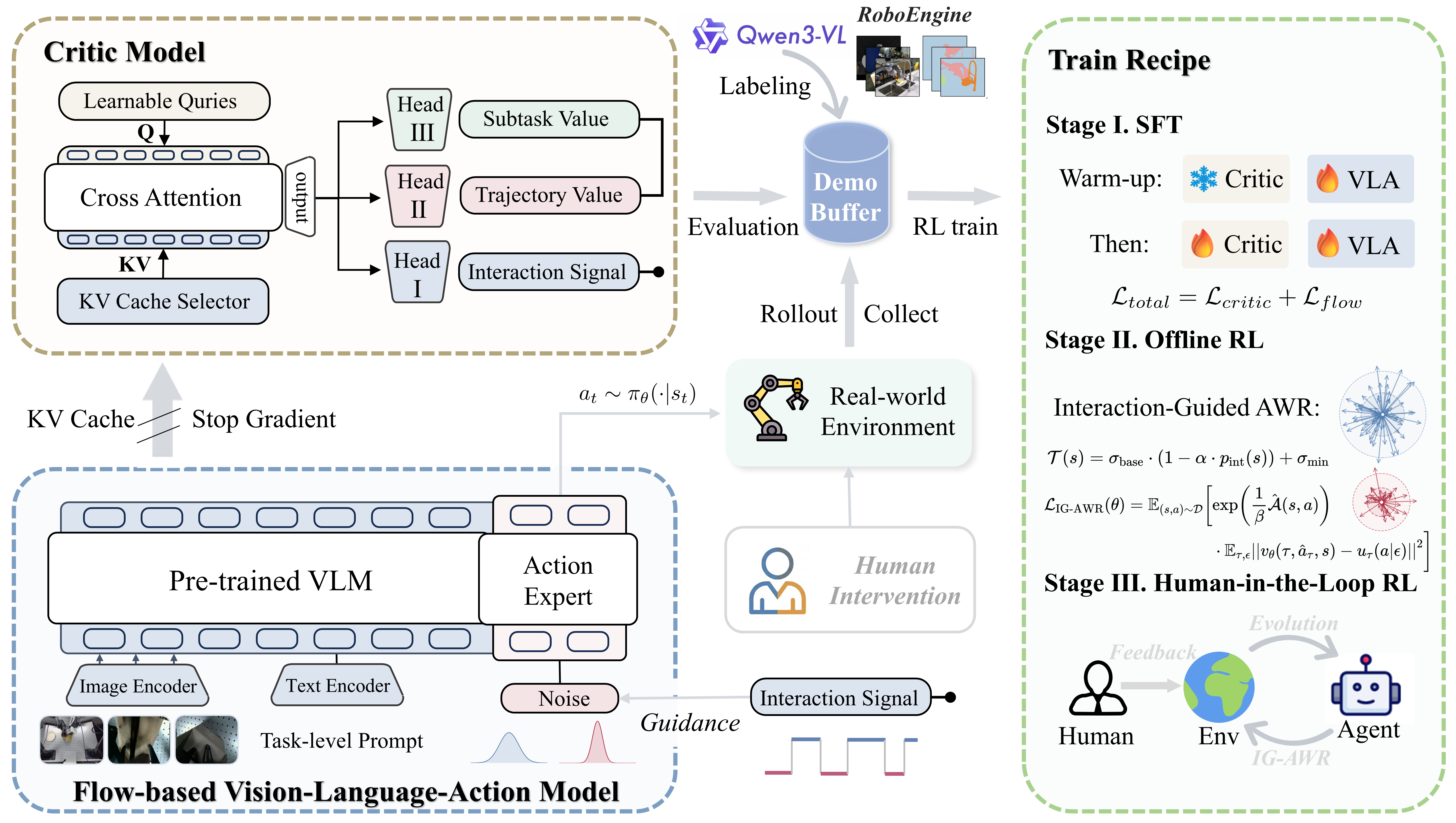}
    \caption{\textbf{Overview of the Interaction-Guided Reinforced Fine-tuning System (IG-RFT).} The system integrates a multi-head Q-Former-based critic model to provide dense value and interaction guidance. The supervision target of the dense value is our designed general hybrid value function. We use the interaction signal to modulate the flow matching initial noise to balance RL exploration and exploitation. The training process progresses through three stages: Supervised Fine-Tuning (SFT), data-efficient Offline RL using the proposed IG-AWR algorithm, and final Human-in-the-Loop refinement to master long-horizon real-world manipulation tasks.}
    \label{fig:overview}
\end{figure*}

In this work, we propose a novel reinforced fine-tuning system for flow-based VLA models, designed to master long-horizon complex robotic manipulation tasks in the real world. As illustrated in Figure~\ref{fig:overview}, the system consists of three stages: SFT, Offline RL, and Human-in-the-Loop RL, enabling a gradual transition from initial policy learning to policy evolution. Our core insight is that effective policy optimization necessitates structured interaction guidance and general dense procedural rewards. To this end, we firstly propose Interaction-Guided Advantage Weighted Regression (IG-AWR), an algorithm that dynamically modulates exploration intensity based on the robot's interaction status. Specifically, this enables the robot to prioritize exploration during non-interaction phases (e.g., approach, retraction) for adapting to the diversity of broadly distributed scenarios, while emphasizing robustness during fine-grained interaction phases (e.g., handover, grasping) to acquire complex manipulation capabilities. To annotate the interaction status for episodes, we leverage open-source robotic segmentation models and optical flow estimation methods for automated processing. Furthermore, we design a novel dense reward function that integrates the trajectory-level reward and the subtask-level reward, providing comprehensive procedural guidance for the RL system, which depends solely on episode lengths and is thus broadly applicable across diverse tasks. While some prior works design dense rewards for specific tasks using visual similarity or distance to target points \cite{zhang2025reinbot, hung2025nora, zhang2024grape, zhai2025vision,zhang2025balancing}, others resort to sparse rewards in real-world reinforcement learning settings \cite{luo2025precise, chen2025conrft, luo2024serl, physical2025pi, li2025gr}. Finally, we design a multi-stage training system that achieves stable and efficient policy learning during the SFT and Offline RL stages, and leverages Human-in-the-Loop RL for iterative policy refinement. The innovation and effectiveness of the proposed components are rigorously validated through extensive experiments, including ablation studies on individual modules and comprehensive evaluation of the entire fine-tuning system on real-world long-horizon complex manipulation tasks.
Our main contributions are summarized as follows:
\begin{itemize}[leftmargin=*]
\item We propose Interaction-Guided AWR (IG-AWR), a novel RL fine-tuning algorithm designed for flow-based VLA models that effectively balances exploration and exploitation by utilizing interaction signals.
\item We formulate a hybrid dense reward function incorporating trajectory-level procedural rewards and weighted subtask-level rewards, delivering smooth and dense feedback for long-horizon tasks.
\item We design a multi-stage reinforced fine-tuning system integrated with human intervention. The system initializes policies via SFT, accomplishes foundational policy learning through Offline RL, and further elevates policy performance during Human-in-the-Loop rollouts.
\end{itemize}

\section{Related Work}
\subsection{Vision-Language-Action Models}
Robotic policies serve as agents that map perceived states to control actions\cite{zhao2023learning,chi2025diffusion}. Vision-Language-Action (VLA) models are generalizable robotic policies equipped with semantic understanding, trained by adapting vision-language models (VLMs) on robotic manipulation demonstrations and open-world datasets\cite{black2410pi0, gr00tn1_2025, black2025pi_, shukor2025smolvla, zheng2025x, ghasemipour2025self, kim2025fine, kim2024openvla,brohan2022rt,zitkovich2023rt}. RT-1 \cite{brohan2022rt} and RT-2 \cite{zitkovich2023rt} are early representative methods that employ transformer-based architectures to predict discretized actions. In contrast, $\pi_0$~\cite{black2410pi0} utilizes an action expert, independent of the pre-trained VLM parameters, to generate actions through flow matching. $\pi_{0.5}$ \cite{black2025pi_} is trained with knowledge isolation \cite{pertsch2025fast} and enables subtask decomposition \cite{shi2025hi},achieving broad generalization in real-world manipulation tasks. 
Despite the rapid recent advancements and demonstrated potential of VLA models, they still exhibit limited generalization and robustness when deployed in novel real-world domains, leading to degraded performance in out-of-distribution scenarios. Advancing fine-tuning methodologies can significantly enhance their capabilities in tackling long-horizon complex real-world tasks.

\subsection{RL Fine-tuning for VLA Models}
Fine-tuning VLA models via Reinforcement Learning (RL) has recently emerged as a prominent research focus. One way is to conduct online RL within simulation using large-scale parallel environments for model training \cite{lu2025vla,liu2025can,yu2025rlinf,chebotar2023q,li2025simplevla, zhang2025gorl}. Some studies target RL training for flow-based VLA models\cite{chen2025pi_,liu2025flow,lyu2025reinforcement,frans2025diffusion,wagenmaker2025steering}; notably, flow-based VLA models present greater challenges for RL optimization compared to auto-regressive VLA models. The world model, serving as a more efficient interaction environment than the physical world, has been explored to provide feedback for RL fine-tuning \cite{hung2025nora,zhang2025reinforcing,zhu2025wmpo}. Reward function design has also been increasingly investigated \cite{fei2025srpo,zhang2024grape,zhang2025reinbot,zhai2025vision,xu2025stare, dai2025rover, chen2025sarm}, as sparse rewards often lead to unstable training, while dense rewards remain difficult to generalize. Another category of methods concentrates on exploring safe and efficient RL fine-tuning in real-world scenarios, mainly through specially devised offline-to-online RL approaches or human-in-the-loop guidance to mitigate the distribution shift problem \cite{physical2025pi, chen2025conrft, xu2024rldg,lu2025human,luo2025precise,li2025gr}; however, these methods lack stage-aware dense reward feedback. In contrast, our work designs a more efficient interaction-guided AWR algorithm, alongside a general and dense reward mechanism. It improves the performance of the system spanning from Offline RL to Human-in-the-Loop RL.

\section{Preliminaries}
\subsection{Vision-Language-Action Models}
Consider a robotic manipulation task specified by a natural language instruction $\ell$. At each time step $t$, the agent receives an observation $o_t$ (proprioception and visual data). VLA models leverage large-scale pre-trained VLMs and an action module as an end-to-end policy, sampling actions $a_t \sim \pi(\cdot|\ell, o_t)$ conditioned on the multimodal context. Standard post-training relies on SFT via objectives like next-token prediction or conditional flow matching, which is highly sensitive to demonstration quality and suffers from data inefficiency. Crucially, SFT indiscriminately treats sub-optimal and optimal demonstrations as equivalent supervision targets.

\subsection{Reinforcement Learning}
The robotic manipulation task is formulated as a Markov Decision Process (MDP) defined by the tuple $\mathcal{M} = (\mathcal{S}, \mathcal{A}, r, \mathcal{P}, \gamma)$. The state $s_t \in \mathcal{S}$ encompasses the robot observation $o_t$ and task prompt $\ell$. The VLA serves as the policy $\pi_\theta(\cdot|s_t)$ to sample actions $a_t$ with the transition dynamics $\mathcal{P}(s_{t+1}|s_t, a_t)$ determining the next state. We define the cumulative return as $G_t = \sum_{k=t}^{T} \gamma^{k-t} r_k$, where $\gamma \in [0, 1]$ is the discount factor. Let $\tau$ denote a trajectory sampled from the policy; the goal of RL is to optimize parameters $\theta$ to maximize the expected return $\mathcal{J}(\theta) = \mathbb{E}_{\tau \sim \pi_\theta} \left[ \sum_{t=0}^{T} \gamma^t r_t \right]$. Accordingly, the state value function is defined as $V^\pi(s_t) = \mathbb{E}_{\tau \sim \pi_\theta} [G_t | s_t]$.

\section{Methodology}
\subsection{Interaction-guided AWR for Flow Matching}
In complex long-horizon robotic manipulation tasks, achieving an optimal balance between exploration and exploitation remains a critical yet unresolved challenge. Conventional RL methods often suffer from detrimental exploration during delicate operations. To address this, we decompose episodes into non-interactive phases (e.g., approach, retraction), where we encourage exploration for trajectory diversity, and contact-rich interaction phases (e.g., handover, grasping), where we prioritize precision and stability. Specifically, we first introduce the extraction of interaction signals, followed by their application to the RL fine-tuning of flow-based VLA models.

\begin{figure}[t]
    \centering
    \includegraphics[width=\columnwidth]{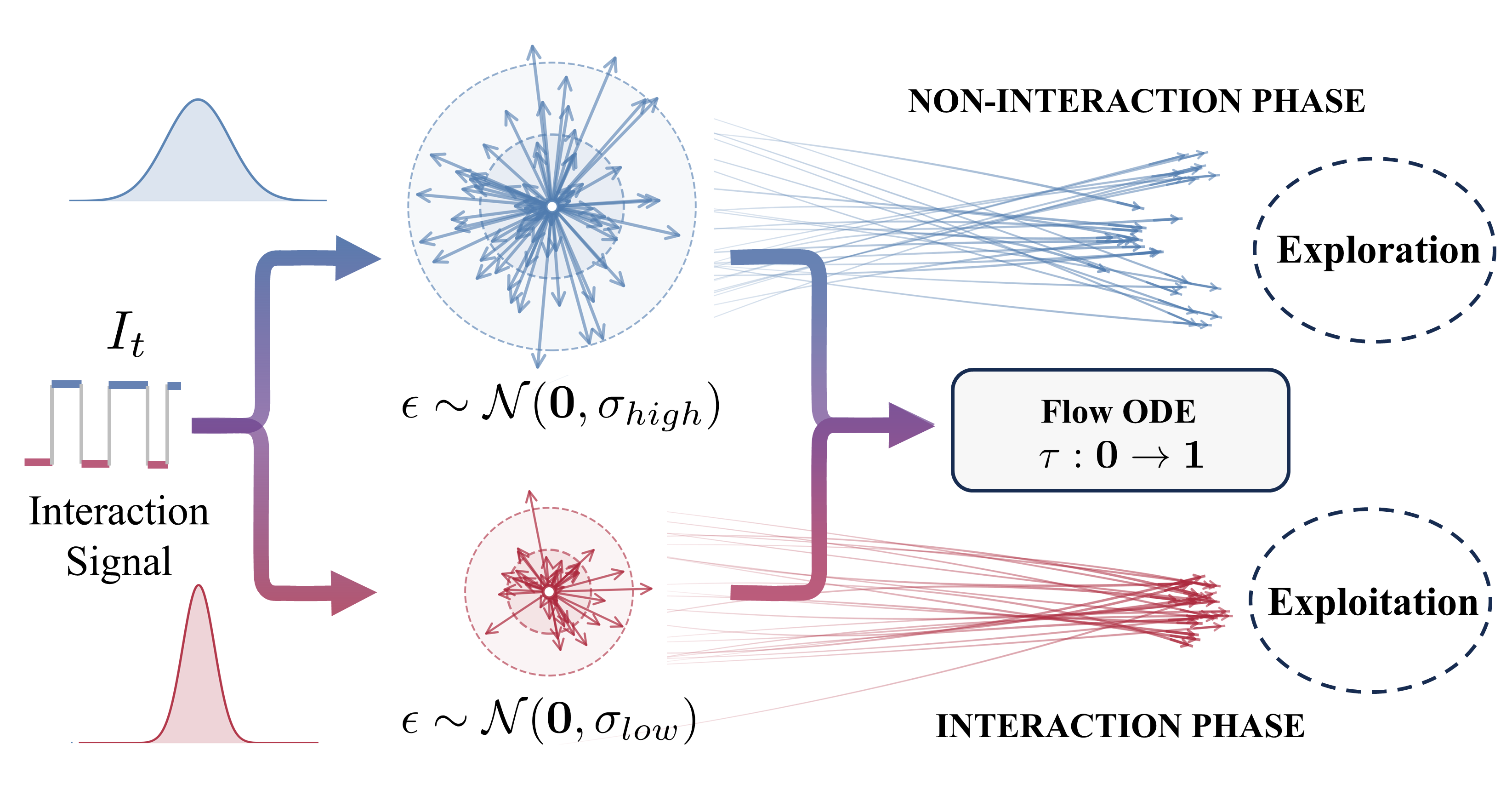}
    \caption{\textbf{Illustration of Dynamic Uncertainty Modulation in IG-AWR.} This mechanism modulates the initial sampling noise $\epsilon$ of the flow ODE based on the extracted interaction signal $I_t$. During non-interaction phases (top), a higher variance $\sigma_{\text{high}}$ is applied to encourage diverse trajectory exploration. Conversely, during interaction phases (bottom), the variance is reduced to $\sigma_{\text{low}}$ to ensure precision and stability for contact-rich manipulation.}
    \label{fig:IG_flow}
\end{figure}

\subsubsection{Interaction Signal Extraction}
\label{sec:interaction_extraction}
Efficient and autonomous interaction labeling is crucial for processing VLA demonstrations. Leveraging global visual observations, at each time step $t$, we first employ the robot segmentation model, RoboEngine\cite{yuan2025roboengine}, to generate a binary robot mask $M_t$. Subsequently, we apply RAFT\cite{teed2020raft} to estimate the dense optical flow field $F_t$. The interaction status $I_t$ is determined by
\begin{equation} I_t = \mathbb{I} \left( \sum_{u,v} (1 - M_{t}^{(u,v)}) \cdot \mathbb{I}\left( || F_{t}^{(u,v)} ||^2 > \delta_{\text{flow}} \right) > \Theta_{\text{pixel}} \right) \end{equation}
where $(u,v)$ denotes pixel coordinates and $\mathbb{I}(\cdot)$ represents the indicator function. The term $(1 - M_{t}^{(u,v)})$ excludes robot pixels to focus on environmental motion. We count the pixels where the squared flow magnitude $|| F_{t}^{(u,v)} ||^2$ exceeds the threshold $\delta_{\text{flow}}$. If this count exceeds $\Theta_{\text{pixel}}$, the time step is labeled as an interaction phase ($I_t=1$).

\subsubsection{Dynamic Uncertainty Modulation}
During the rollout phase, as illustrated in Figure~\ref{fig:IG_flow}, we dynamically regulate the variance of the initial sampling noise conditioned on the predicted interaction signal probability. Building directly on the binary interaction labels derived from visual observations (Section \ref{sec:interaction_extraction}), the critic network predicts 
 $p_{\text{int}}(s) \in [0,1]$—a soft estimate of interaction likelihood for the state, trained to approximate the ground truth. We define the effective sampling temperature $\mathcal{T}(s)$ as:\begin{equation}\mathcal{T}(s) = \sigma_{\text{base}} \cdot (1 - \alpha \cdot p_{\text{int}}(s)) + \sigma_{\text{min}}\end{equation}The policy generates actions by solving the flow ODE starting from a scaled noise distribution. Specifically, the initial state $\hat{a}_0$, equal to $\epsilon$, is sampled from $\mathcal{N}(\mathbf{0}, \mathcal{T}^2(s)\mathbf{I})$.

This formulation adaptively modulates exploration: high variance during non-interaction phases encourages diverse trajectory exploration, while low variance during interaction phases enforces precise, stable execution—seamlessly bridging the discrete visual-derived interaction signal $I_t$ with continuous policy uncertainty modulation.

\subsubsection{Advantage-Weighted Flow Update}
For policy optimization, we employ Advantage Weighted Regression (AWR) \cite{peng2019advantage} to maximize the expected return of the flow-based policy. Given a demonstration $\mathcal{D}$ collected via the interaction-guided policy, we estimate the advantage $\hat{\mathcal{A}}(s_t, a_t) = r_{t:t+N}+V_\phi(s_{t+N}) - V_\phi(s_t)$. The Interaction-Guided Advantage Weight Regression (IG-AWR) loss function is formulated as:
\begin{equation}
\begin{split}
    \mathcal{L}_{\text{IG-AWR}}(\theta) = \mathbb{E}_{(s, a) \sim \mathcal{D}} \bigg[ & \exp\left(\frac{1}{\beta}\hat{\mathcal{A}}(s, a)\right) \\
    & \cdot \mathbb{E}_{\tau, \epsilon} \left|\left| v_\theta(\tau, \hat{a}_{\tau}, s) - u_{\tau}(a|\epsilon) \right|\right|^2 \bigg]
\end{split}
\end{equation}
where $\beta$ is the temperature hyperparameter for AWR, and $\hat{a}_{\tau}$ denotes the flow trajectory at time $\tau$. This objective upweights data with higher estimated advantages while preserving the flow ODE structure for stable policy refinement.

\begin{figure*}[t]
    \centering
    \includegraphics[width=2\columnwidth]{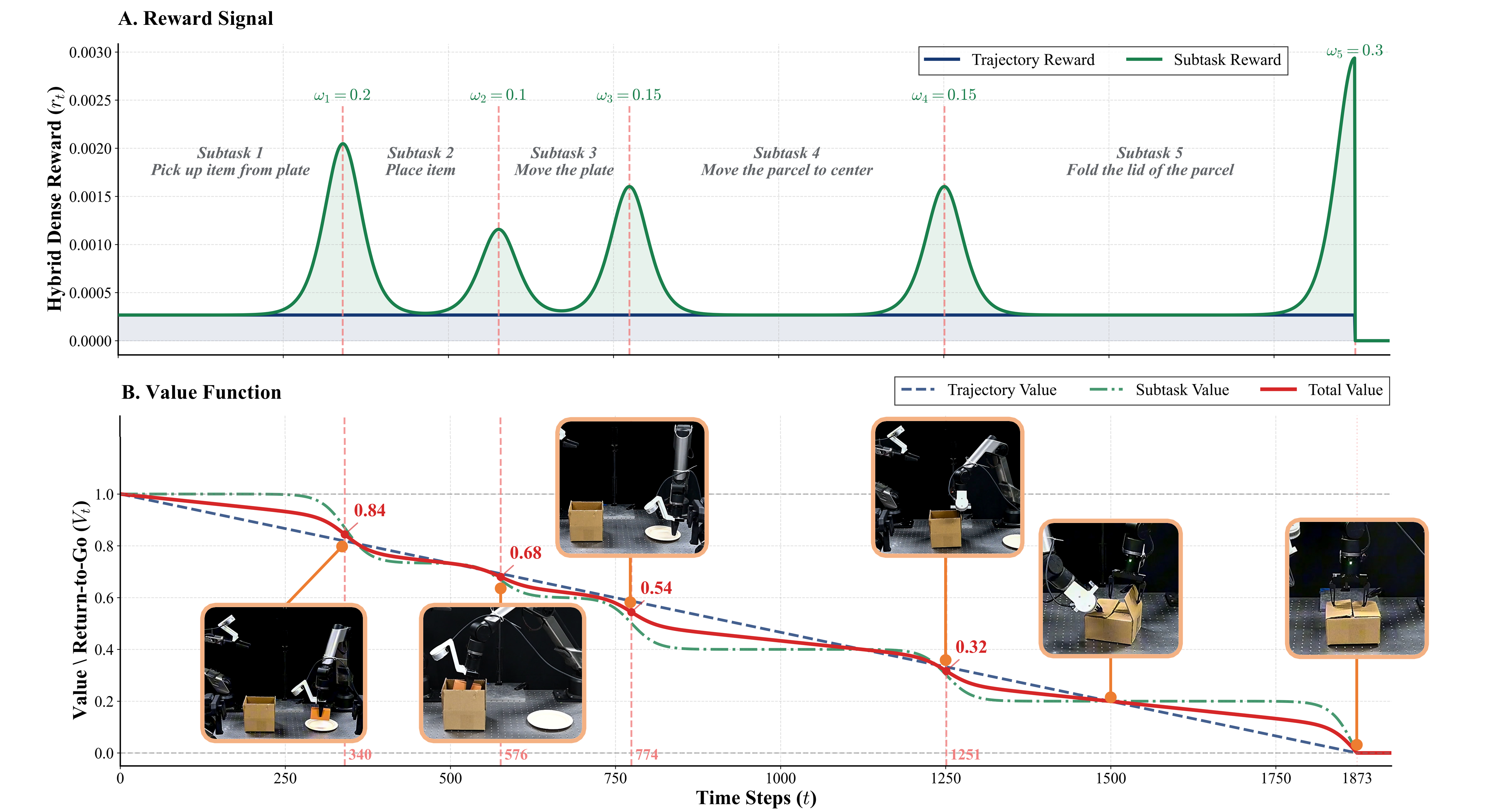}
    \caption{\textbf{Visualization of the Reward Signal and Value Function.} We illustrate the proposed reward shaping mechanism on the "Parcel Packing" task. \textbf{(A)} The reward signal $r_t$ combines a constant trajectory-level reward with weighted subtask-level rewards (peaks), providing dense feedback at key milestones. \textbf{(B)} The Value (Return-to-Go) function $V_t$ monotonically decreases from 1.0 to 0 as the agent progresses through subtasks, serving as a precise progress indicator for the Critic.}
    \label{fig:reward_function}
\end{figure*}

\subsection{General Hybrid Reward Function}
Reward modeling is critical for RL fine-tuning of VLA models. Designing a general reward function is beneficial for enabling adaptation to a wide range of tasks. To effectively guide the agent through multi-stage long-horizon tasks, we design a hybrid reward function that incorporates both trajectory-level success incentives and stage-wise subtask feedback, as illustrated in Figure~\ref{fig:reward_function}

\subsubsection{Trajectory-level Reward}
To generalize across various robotic manipulation tasks, we define the trajectory-level reward as:
\begin{equation}
    r_t^{\text{traj}} =
    \begin{cases}
        1/T & \text{if task succeeds} \\
        0 & \text{otherwise}
    \end{cases}
\end{equation}
where $T$ denotes the total number of time steps in the episode. This formulation distributes the total return of $1.0$ uniformly across the trajectory upon successful task completion.

\subsubsection{Subtask-level Reward}
Since the trajectory-level reward lacks fine-grained feedback for complex long-horizon tasks, we employ potential-based reward shaping at the subtask level. This continuous monotonically increasing potential function prevents reward hacking and provides sufficient local guidance. To simulate subtask completion while ensuring smoothness, we adopt a weighted mixture model to prioritize complex subtasks, formulated as:
\begin{align}
    \Phi(\rho) &= C \sum_{i=1}^K w_i \cdot \sigma\left(\frac{\rho - k_i}{\epsilon}\right) \quad \text{s.t.} \sum w_i = 1\\
    r_t^{\text{sub}} &= \Phi(\rho_t) - \Phi(\rho_{t-1})
\end{align}
where $\rho_t = t/T$ denotes the normalized temporal progress, and $\sigma(\cdot)$ is the standard sigmoid function. The parameters $k_i$ and $w_i$ represent the completion time steps and the importance weight of the $i$-th subtask. Here, $\epsilon$ controls the sharpness and $C$ is a normalization constant.

\subsubsection{Reward Aggregation}
We obtain the final reward by computing a weighted sum of the trajectory-level reward and the subtask-level reward:
\begin{equation}
r_t = \omega_\text{traj}\cdot r_t^{\text{traj}} + \omega_\text{sub} \cdot r_t^{\text{sub}} \quad \text{s.t.} \quad\omega_\text{traj}+\omega_\text{sub}=1
\end{equation}
Accordingly, we derive the return-to-go as $G_t = \sum_{k=t}^{T} r_k$ (assuming a discount factor $\gamma=1$). This value naturally decays from $1.0$ to $0$ as the episode progresses.

\subsection{RL Fine-tuning System}
Due to low data efficiency in purely online RL, we deploy a multi-stage fine-tuning method in which offline training with pre-collected data provides a stronger prior. In the online stage, human feedback provides high-quality and efficient guidance for RL fine-tuning, mitigating exploration challenges in high-dimensional spaces. Building on this paradigm, we design a system incorporating the IG-AWR algorithm and dense reward function for long-horizon complex tasks.

\subsubsection{Model Architecture}
We adopt the $\pi_{0.5}$ model \cite{black2025pi_}, a representative flow-based VLA validated for superior real-world performance, as our actor module $\pi_\theta$. Leveraging the KV cache from the VLM backbone conditioned on the current observation $o_t$, we train a critic network to regress the state value $V^\pi_t$ and predict the interaction signal $I_t$. Specifically, utilizing the KV cache from all transformer layers incurs computational inefficiency and information redundancy. Therefore, we selectively aggregate features from the deeper layers, which contain the richest semantic information. The network architecture is based on the Q-Former (Query Transformer) \cite{li2023blip}, which utilizes learnable queries to adaptively extract relevant multimodal features from the KV cache through cross-attention. The Q-Former output is then passed through three separate parameterized heads to produce the trajectory-level value, subtask-level value, and interaction signal, respectively. The loss of the Critic Module should not affect action generation; thus, we stop gradient propagation to the VLM. Further details on the critic network architecture are provided in Appendix~\ref{sec:critic_architecture}.

\begin{algorithm}[t]
\caption{Interaction-Guided AWR for VLA Fine-tuning}
\label{alg:hitl_ig_awr}
\begin{algorithmic}[1] 
    \Require Pre-trained VLA $\pi_{0.5}$, Demonstration Buffer $\mathcal{D}_{\text{demo}}$
    \Ensure Fine-tuned Policy $\pi_\theta$, Critic $V_\phi$

    \State \textbf{Data Annotation:} Annotate $\mathcal{D}_{\text{demo}}$ with subtasks, interaction signals $I_t$, rewards, and values.

    \State \textcolor{gray}{\textit{// Stage I: Warm-up and SFT}}
    \State Initialize Actor $\pi_\theta$ from $\pi_{0.5}$ via SFT.
    \State Initialize Critic $V_\phi$ by minimizing $\mathcal{L}_{\text{MSE}}$ (State Value) and $\mathcal{L}_{\text{BCE}}$ (Interaction Signal) on $\mathcal{D}_{\text{demo}}$.

    \State \textcolor{gray}{\textit{// Stage II: Offline RL}}
    \For{iteration $k = 1, \dots, K_{\text{offline}}$}
        \State Estimate Advantage $\hat{\mathcal{A}}$ on $\mathcal{D}_{\text{demo}}$.
        \State Update $\pi_\theta$ via IG-AWR and update $V_\phi$.
    \EndFor

    \State \textcolor{gray}{\textit{// Stage III: Human-in-the-Loop RL}}
    \State Initialize Replay Buffer $\mathcal{D}_{\text{real}} \leftarrow \emptyset$
    \While{not converged}
        \State $s_0 \leftarrow \text{env.reset()}$
        \For{$t = 0, \dots, T$}
            \State Observe state $s_t$, Critic outputs $V(s_t), I(s_t)$.
            \If{$V(s_t)$ stagnates \textbf{or} Human identifies failure}
                \State $a_t \leftarrow \text{HumanTeleop}(s_t)$ \Comment{Human Override}
            \Else
                \State $a_t \sim \pi_\theta(\cdot|s_t)$ \Comment{Autonomous Execution}
            \EndIf
            \State Execute $a_t$, observe $s_{t+1}$.
        \EndFor
        \State Store episode to $\mathcal{D}_{\text{real}}$.

        \If{Update Interval}
            \State Sample batch $\mathcal{B}_{\text{real}} \sim \mathcal{D}_{\text{real}}$.
            \State Sample equal-sized batch $\mathcal{B}_{\text{demo}} \sim \mathcal{D}_{\text{demo}}$.
            \State Construct Hybrid Dataset $\mathcal{B}_{\text{hybrid}} = \mathcal{B}_{\text{real}} \cup \mathcal{B}_{\text{demo}}$.
            \State Update Policy $\pi_\theta$ via IG-AWR on $\mathcal{B}_{\text{hybrid}}$.
            \State Update Critic $V_\phi$ (at a lower frequency).
        \EndIf
    \EndWhile
\end{algorithmic}
\end{algorithm}

\subsubsection{Training Recipe}
With the proposed architecture established, we now outline the optimization procedure. After obtaining the state value function, we estimate the advantage $\hat{\mathcal{A}}(s_t, a_t)$ using N-step advantage estimation \cite{physical2025pi} to update the policy via the IG-AWR algorithm. Specifically, we devise a three-stage RL training curriculum for the VLA models, as summarized in Algorithm~\ref{alg:hitl_ig_awr}:
\begin{itemize}[leftmargin=*]
\item \textbf{Stage I: Warm-up and SFT.} We collect demonstration data and employ a large-scale VLM to annotate subtask information. Interaction signals, rewards, and values are obtained. We use Knowledge Isolation \cite{driess2025knowledge} to train $\pi_{0.5}$ (including subtask prediction loss). After a few warm-up steps, we train the critic module simultaneously using MSE loss for value regression and BCE loss for interaction prediction.
\item \textbf{Stage II: Offline RL.} We implement the IG-AWR algorithm to conduct RL training for the VLA and simultaneously update the critic module. In this phase, the agent learns to evaluate and utilize data, achieving efficient policy learning.
\item \textbf{Stage III: Human-in-the-Loop RL.} We deploy the base policy in the real world to collect rollouts with Human-in-the-Loop intervention. By collecting high-quality human corrections, we enable the policy to approach its capability upper bound, realizing performance improvements especially for long-horizon complex tasks.
\end{itemize}

\subsubsection{Human-in-the-Loop Mechanism}
During real-world execution, the robot operates autonomously according to the policy $\pi_\theta$ while a human operator monitors the process. When the value output by the critic module exhibits stagnation or the human identifies a potential failure, the operator intervenes via a teleoperation interface to override the action. These intervention data are stored in the replay buffer, capturing both the low-value states preceding the takeover and the subsequent high-quality corrective trajectories. Periodically, we sample an equal number of episodes from the pre-collected demonstration buffer to construct a hybrid dataset. We then fine-tune the policy module using this dataset via the IG-AWR algorithm, while the critic module is updated at a lower frequency. This Human-in-the-Loop RL approach effectively overcomes distribution shifts and enables the mastery of complex long-horizon tasks that are difficult to solve via offline learning.

\section{Experiments}
In this section, we conduct comprehensive real-world experiments to evaluate the effectiveness of our proposed reinforced fine-tuning system. Our experiments are designed to answer three key research questions:
\begin{itemize}
    \item \textbf{Q1 (Overall Performance):} Does our reinforced fine-tuning system perform robustly in long-horizon complex real-world tasks and outperform SFT/RL baselines?
    \item \textbf{Q2 (Ablation Study):} What are the contributions of individual components, specifically the IG-AWR algorithm and the hybrid dense reward function?
    \item \textbf{Q3 (System Evolution):} How does the policy performance evolve across the three training stages (SFT, Offline RL, and Human-in-the-Loop RL)?
\end{itemize}

\subsection{Experimental Setup}
\subsubsection{Hardware and Environment}
We perform all experiments on a real-world robotic platform consisting of a Galaxea A1 dual-arm robot equipped with parallel-jaw grippers. Visual observations are captured via a head-mounted Orbbec Gemini 2L camera and two wrist-mounted RealSense L515 cameras. The control frequency is set to 30 Hz, matching the data collection frequency. We set the action chunk size to 30 and employ the RTC method \cite{black2025real} for real-time control. Policy inference and training are executed on a cloud server equipped with an NVIDIA A100 GPU. The low-level control pipeline runs on a local PC with an Intel Core i7-12700KF 3.6GHz CPU, 32GB RAM, and an RTX 4070 GPU. Socket-based network communication is established between the local deployment endpoint and the cloud server inference endpoint. For detailed training hyperparameters and configurations, please refer to Appendix~\ref{sec:hyperparams}.

\subsubsection{Tasks}
\begin{figure*}[t]
    \centering
    \includegraphics[width=2\columnwidth]{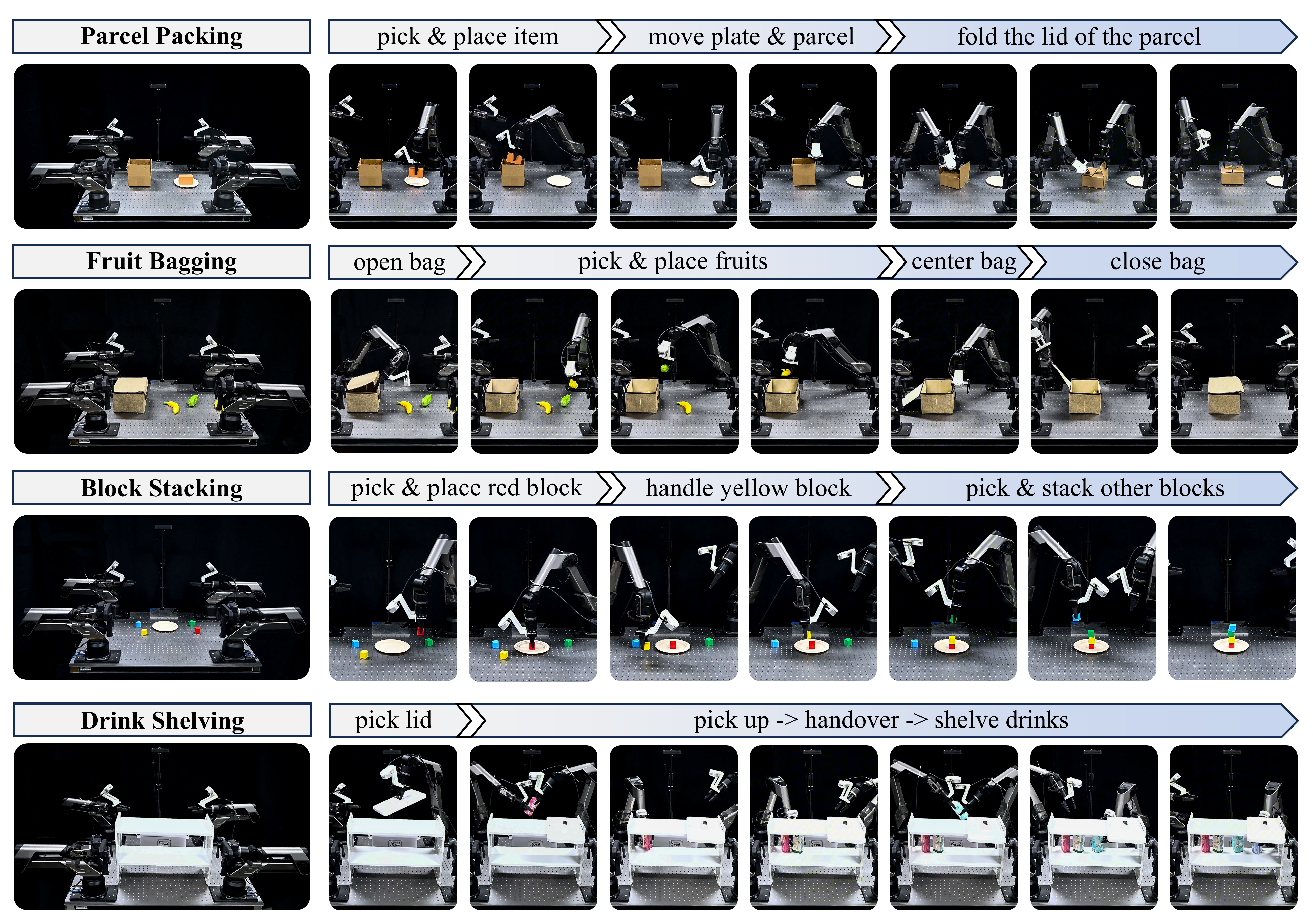}
    \caption{\textbf{Execution Sequences of Long-Horizon Tasks.} From top to bottom, we visualize the successful execution flows for four challenging tasks: Parcel Packing, Fruit Bagging, Block Stacking, and Drink Shelving. Arrows indicate the progression of subgoals. \textbf{Note:} The text labels above each stage are abbreviated for visualization clarity and differ from the actual natural language instructions used by the model. For the detailed sequence of subtask instructions for each task, please refer to Appendix~\ref{sec:tasks}.}
    \label{fig:experiment_tasks}
\end{figure*}

We select four challenging long-horizon manipulation tasks that involve multi-stage reasoning, contact-rich interactions, bimanual coordination, as well as articulated and deformable object manipulation. Task summaries are provided below and visualized in Figure~\ref{fig:experiment_tasks}; comprehensive descriptions—including subtask decomposition, success criteria and average steps—are available in Appendix~\ref{sec:tasks}.
\begin{itemize}
    \item \textbf{Parcel Packing:} The robot must pick up an item on the plate, place it into the parcel, move the plate aside, and perform dual-arm collaboration to synchronously fold the flaps and close the box. This task requires precise bimanual coordination and contact-rich manipulation capabilities.
    \item \textbf{Fruit Bagging:} The agent is required to open the lid of the canvas bag, sequentially pick up ginger, bitter melon, and banana, and place them into the bag. Finally, it needs to close the lid. This task involves articulated and deformable object manipulation, semantic understanding of object categories, and sequential picking from random positions.
    \item \textbf{Block Stacking:} The robot needs to pick up colored blocks from the table in the order of red, yellow, green, and blue, and sequentially stack them onto a plate. This task tests the capability for color-semantic understanding and precise stacking.
    \item \textbf{Drink Shelving:} The robot is required to retrieve beverages from a cluttered container, where the closer arm picks items from the top and performs a handover to the other arm, which then places them into narrow shelves from the side. This task tests dual-arm collaborative handover, bin-picking in cluttered environments, and precise placement in confined shelf spaces.
\end{itemize}

\begin{table}[t]
\caption{\textbf{Main Results on Real-world Long-horizon Tasks.} We compare the success rate (\%) of our system against baselines across four tasks. Results are averaged over 20 trials per task. Best results are highlighted in \textbf{BOLD}.}
\label{tab:main_results}
\centering
\resizebox{\linewidth}{!}{
\begin{tabular}{lccccc}
\toprule
\textbf{Method} & \textbf{\shortstack{Parcel\\Packing}} & \textbf{\shortstack{Fruit\\Bagging}} & \textbf{\shortstack{Block\\Stacking}} & \textbf{\shortstack{Drink\\Shelving}} & \textbf{Avg.} \\
\midrule
SFT & 15.0 & 30.0 & 25.0 & 5.0 & 18.8 \\
IQL & 40.0 & 50.0 & 45.0 & 20.0 & 38.8 \\
AWR & 35.0 & 55.0 & 45.0 & 25.0 & 40.0 \\
\textbf{Ours} & \textbf{85.0} & \textbf{95.0} & \textbf{90.0} & \textbf{70.0} & \textbf{85.0} \\
\bottomrule
\end{tabular}
}
\end{table}

\subsubsection{Baselines}
We compare our full system against the following baselines:
\begin{itemize}
    \item \textbf{SFT ($\boldsymbol{\pi_{0.5}}$-Base):} The base flow-based VLA model fine-tuned solely via behavior cloning on the demonstration dataset. Action generation is trained using Knowledge Isolation \cite{driess2025knowledge} while subtask prediction is performed concurrently.
    \item \textbf{Offline RL (IQL/AWR):} Standard Offline RL methods~\cite{kostrikov2021offline, peng2019advantage} applied to the $\pi_{0.5}$ model, lacking our specific interaction guidance, dense reward shaping, and human intervention.
\end{itemize}

\subsubsection{Data Collection and Metrics}
For each task, we collect 60 expert demonstrations via teleoperation to form the initial dataset $\mathcal{D}_{\text{demo}}$. During the Human-in-the-Loop stage, we perform 10 rollouts per training update and add them to $\mathcal{D}_{\text{real}}$ for IG-AWR updates, iterating 4 times (totaling 40 rollouts). For fair comparison, baseline methods without Human-in-the-Loop utilize 100 expert demonstrations. The metric is \textbf{Success Rate (SR)}, defined as the ratio of successful trials to the total number of attempts.

\subsection{Main Results}
We present the comparative results in Table~\ref{tab:main_results}. Our proposed system achieves the highest success rates across all four tasks, demonstrating superior generalization and robustness.

\subsubsection{Superiority over SFT}
Although VLA models learn diverse manipulation skills and semantic understanding through SFT during the pre-training stage, becoming a generalist policy, current SOTA methods still struggle to transfer knowledge to new real-world domains—including new embodiments and new tasks—using only a few (e.g., 100) task-specific demonstrations to accomplish long-horizon complex tasks. Moreover, SFT is typically constrained by the quality of manually pre-collected data and is more prone to converge to suboptimal distributions. In contrast, RL fine-tuning VLA models for new domains demonstrates superior adaptation. As shown in Table~\ref{tab:main_results}, the SFT policy achieves the lowest success rates in our real-world complex tasks, underperforming all other methods that employ RL fine-tuning. Our method, in particular, significantly improves the success rate over the SFT policy. The results demonstrate that fine-tuning VLA models using RL indeed enables the policy to learn a better action distribution, thereby avoiding compounding errors caused by minor deviations, enhancing the capability to handle long-horizon tasks in new real-world domains.

\subsubsection{Advantage over Offline RL}
Deploying standard Offline RL baselines (IQL/AWR) to fine-tune VLA models can learn a better task-specific action distribution compared to SFT; however, these methods still suffer from distribution shift problems. Meanwhile, purely online RL on real-world embodiments faces challenges of low exploration efficiency and safety concerns. In contrast, the Human-in-the-Loop approach provides human prior guidance to efficiently drive policy evolution. Compared to Offline RL (Table~\ref{tab:main_results}), our multi-stage RL fine-tuning system achieves substantially higher performance gains and demonstrates greater robustness in the real world. In our experimental setting, our method improves performance by approximately 40\% across all four long-horizon complex tasks. This validates the effectiveness and practicality of our reinforced fine-tuning system, thereby answering \textbf{Q1}.

\subsection{Ablation Studies}
To validate our design choices, we conduct ablation studies on the \textbf{Parcel Packing} and \textbf{Drink Shelving} tasks, which demand the most complex manipulation capabilities. This section answers \textbf{Q2}.

\subsubsection{Effect of Interaction-Guided AWR}
We compare our IG-AWR with a standard AWR implementation that also follows the three-stage training recipe but uses uniform exploration noise. As illustrated in Table~\ref{tab:ablation_study}, removing the interaction-guided dynamic uncertainty modulation results in a performance drop of approximately 15\%. Qualitative analysis reveals that without interaction guidance, the robot tends to jitter during the delicate "flap folding and box closing" phase. Furthermore, we observe that without interaction guidance, the agent lacks sufficient exploration during non-interactive phases, leading to poor generalization regarding object positions. It often drifts towards the object locations seen in the pre-collected data, consequently causing task failure when facing out-of-distribution (OOD) cases during testing.

\begin{figure}[t]
    \centering
    \includegraphics[width=\columnwidth]{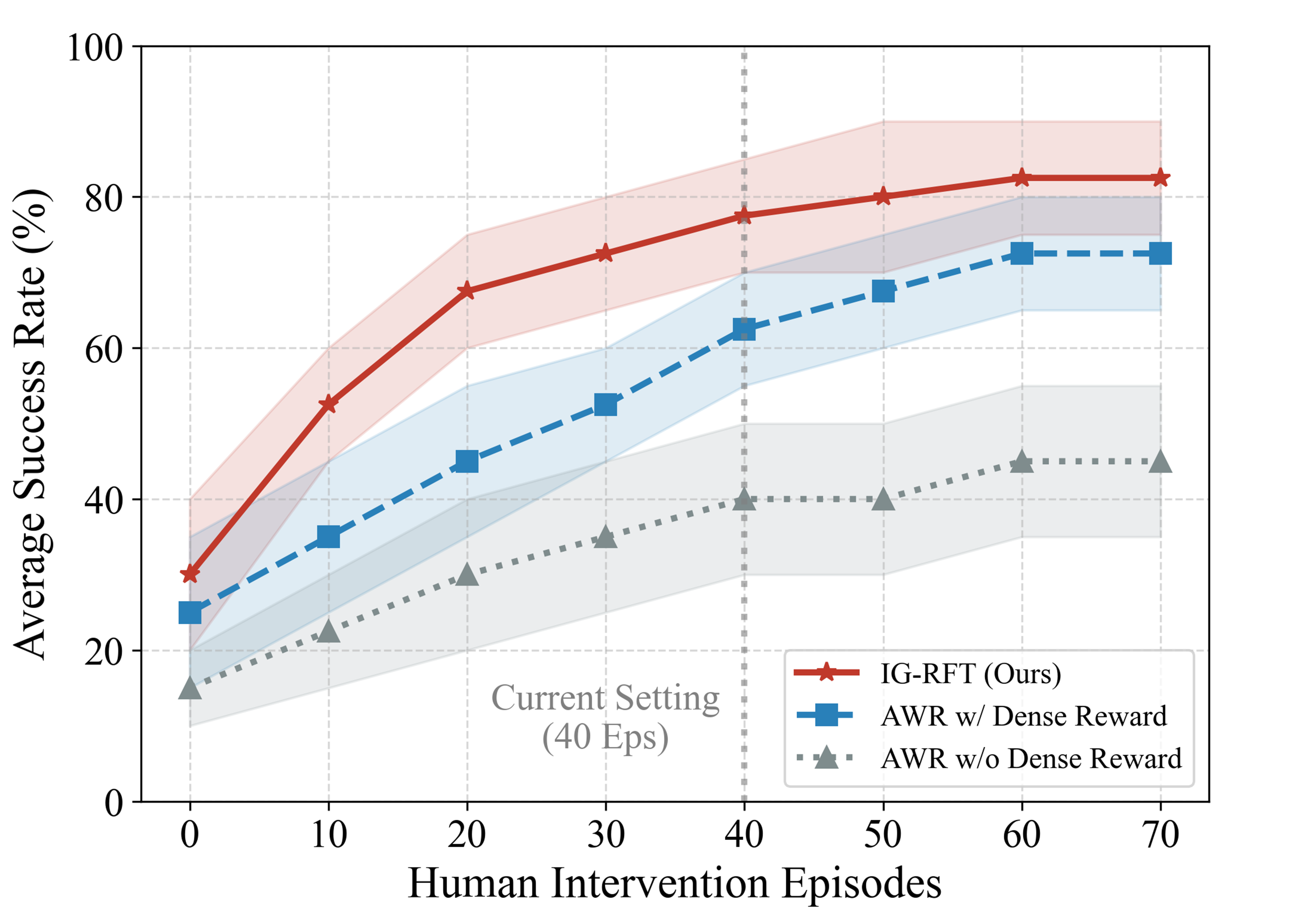}
    \caption{\textbf{Data Efficiency Analysis in Human-in-the-Loop Fine-tuning.} We report the mean success rates averaged over the Parcel Packing and Drink Shelving tasks. The curves show that IG-RFT (Ours) achieves superior sample efficiency, converging to 77.5\% success with only 40 interventions, significantly outperforming baselines. Shaded regions indicate the min-max range across tasks.}
    \label{fig:experiment_ig_ablation}
\end{figure}

Beyond final performance, we further investigate the impact of IG on data efficiency in Figure~\ref{fig:experiment_ig_ablation}. IG-RFT is represented by the red curve, while the baseline AWR w/ dense reward is represented by the blue curve and AWR w/o dense reward by the gray curve. Our method exhibits a significantly steeper learning trajectory. IG-RFT achieves an average success rate of 77.5\% with only 40 human interventions, and ultimately converges to an 82.5\% success rate. The AWR w/ dense reward method learns more slowly, reaching only 62.5\% under the same 40 human interventions. This demonstrates that our method achieves superior performance with significantly lower human cost. Even when the data scale is extended to 70 episodes, the baseline without IG fails to match the peak performance of IG-RFT. This confirms that modulating exploration based on visual interaction signals not only improves policy performance but also drastically reduces the human effort required for fine-tuning. The other baseline, AWR w/o dense reward, further validates this conclusion and highlights that our dense reward is equally significant for policy improvement in Human-in-the-Loop learning.

\begin{table}[t]
\caption{\textbf{Ablation Study Results.} We evaluate the contribution of Interaction Guidance (IG) and Reward Function on the two most challenging tasks.}
\label{tab:ablation_study}
\centering
\resizebox{1.0\linewidth}{!}{
\begin{tabular}{lcc|c}
\toprule
\textbf{Configuration} & \textbf{\shortstack{Parcel\\Packing}} & \textbf{\shortstack{Drink\\Shelving}} & \textbf{Avg. Drop} \\
\midrule
\textbf{Ours (Full System)} & \textbf{85.0} & \textbf{70.0} & - \\
\midrule
\textit{w/o Interaction Guidance} & 70.0 & 55.0 & $\downarrow$ 15.0 \\
\textit{w/o Trajectory-level Reward} & 75.0 & 60.0 & $\downarrow$ 10.0 \\
\textit{w/o Subtask-level Reward} & 55.0 & 45.0 & $\downarrow$ 27.5 \\
\bottomrule
\end{tabular}
}
\end{table}

\begin{table}[t]
\caption{\textbf{Subtask Completion Progress.} Instead of binary success rates, we report the normalized progress score (0.0-1.0), representing the average proportion of subtasks completed per episode. This metric reveals when the policy fails.}
\label{tab:subtask_progress}
\centering
\resizebox{1.0\linewidth}{!}{
\begin{tabular}{lccc}
\toprule
\textbf{Reward Configuration} & \textbf{\shortstack{Parcel\\Packing}} & \textbf{\shortstack{Drink\\Shelving}} & \textbf{Avg. Progress} \\
\midrule
\textbf{Ours (Hybrid)} & \textbf{0.96} & \textbf{0.88} & \textbf{0.92} \\
\midrule
\textit{w/o Trajectory Reward} & 0.84 & 0.78 & 0.81 \\
\textit{w/o Subtask Reward} & 0.68 & 0.54 & 0.61 \\
\bottomrule
\end{tabular}
}
\end{table}

\subsubsection{Effect of Hybrid Dense Reward}
We evaluate the system using only the trajectory-level reward or the subtask-level reward compared to our hybrid dense reward (Trajectory + Subtask) in Table~\ref{tab:ablation_study}, and further analyze failure modes using the subtask completion progress metric in Table~\ref{tab:subtask_progress}. The results indicate that relying on a single reward signal leads to degraded task success rates, which is particularly evident in our long-horizon complex task settings. Specifically, the lack of the subtask-level reward results in a substantial 27\% drop in success rate and a significant 0.31 drop in progress score. This demonstrates that dense incentive signals are critical for facilitating the completion of sequential sub-goals. In contrast, the absence of the trajectory-level reward causes a 10\% drop in success rate, yet the progress score decreases only marginally by 0.11. This indicates that the subtask reward facilitates the completion of intermediate sub-goals, even if the task ultimately fails. We observe that the trajectory-level reward provides long-term guidance towards the final goal, offering macroscopic assurance for the agent to learn the ultimate objective. Conversely, the subtask-level reward offers immediate guidance for short-term objectives, enabling the agent to master sequential sub-goals progressively. Collectively, these results demonstrate that the hybrid reward structure is highly beneficial for the reinforced fine-tuning system.

\begin{table*}[t]
\caption{\textbf{Policy Evolution and Stage Analysis.} We compare the performance at different training stages. Results are averaged over 20 trials per task. \textbf{HIL} denotes Human-in-the-Loop. Best results are highlighted in \textbf{BOLD}.}
\label{tab:training_stages}
\centering
\resizebox{0.95\textwidth}{!}{ 
\begin{tabular}{l|cccc|cc}
\toprule
\textbf{Training Stage / Configuration} & \textbf{Parcel Packing} & \textbf{Fruit Bagging} & \textbf{Block Stacking} & \textbf{Drink Shelving} & \textbf{Avg.} & \textbf{Avg. Drop} \\
\midrule
\textbf{Stage I:} SFT & 15.0 & 30.0 & 25.0 & 5.0 & 18.8 & $\downarrow$ 66.2 \\
\textbf{Stage II:} Offline RL & 40.0 & 55.0 & 45.0 & 20.0 & 40.0 & $\downarrow$ 45.0 \\
\midrule
\textit{SFT $\to$ HIL (w/o Stage II)} & 65.0 & 75.0 & 70.0 & 45.0 & 63.8 & $\downarrow$ 21.2 \\
\textbf{Stage III:} Full System & \textbf{85.0} & \textbf{95.0} & \textbf{90.0} & \textbf{70.0} & \textbf{85.0} & - \\
\bottomrule
\end{tabular}
}
\end{table*}

\subsection{Analysis of Training Stages}
We analyze the policy evolution throughout the three-stage RL training system in Table~\ref{tab:training_stages}, investigating the contribution of each phase to the final performance. This section answers \textbf{Q3}. Detailed analysis for each stage is as follows.

\begin{itemize}
    \item \textbf{Stage I (SFT):} The supervised fine-tuning stage establishes adaptation to the new domain, acquires basic manipulation skills and semantic understanding capabilities, but exhibits limited robustness in long-horizon execution, achieving an average success rate of only 18.8\%. This indicates that the generalization and versatility of VLA models still have substantial room for improvement in real-world settings.
    
    \item \textbf{Stage II (Offline RL):} By optimizing the policy using the interaction-guided objective on pre-collected datasets, the agent learns to distinguish high-value actions from sub-optimal ones, improving the success rate significantly to 40.0\%. Although distribution shift problems remain during real-world rollouts, limiting the success rate, this stage provides a crucial warm-up for the subsequent online phase.
    
    \item \textbf{Stage III (Full System vs. Direct HIL):} Introducing human intervention provides critical corrections for OOD states. As shown in Table~\ref{tab:training_stages}, our full system (Stage I $\to$ II $\to$ III) achieves a superior success rate of \textbf{85.0\%}. Notably, skipping Stage II and performing Human-in-the-Loop directly on the SFT model (\textit{Stage I $\to$ III}) yields a suboptimal success rate of 63.8\%. This performance gap (21.2\%) highlights that the basic distribution of the actor and critic modules learned in Stage II is essential for maximizing the sample efficiency of human corrections. The results show that our three-stage RL fine-tuning system is indispensable for policy evolution.
\end{itemize}

\subsection{Qualitative Analysis}
To provide intuitive insights into the internal mechanisms of IG-RFT, we visualize the extraction of interaction signals and the dynamics of value estimation during real-world execution.

\begin{figure}[t]
    \centering
    \includegraphics[width=\columnwidth]{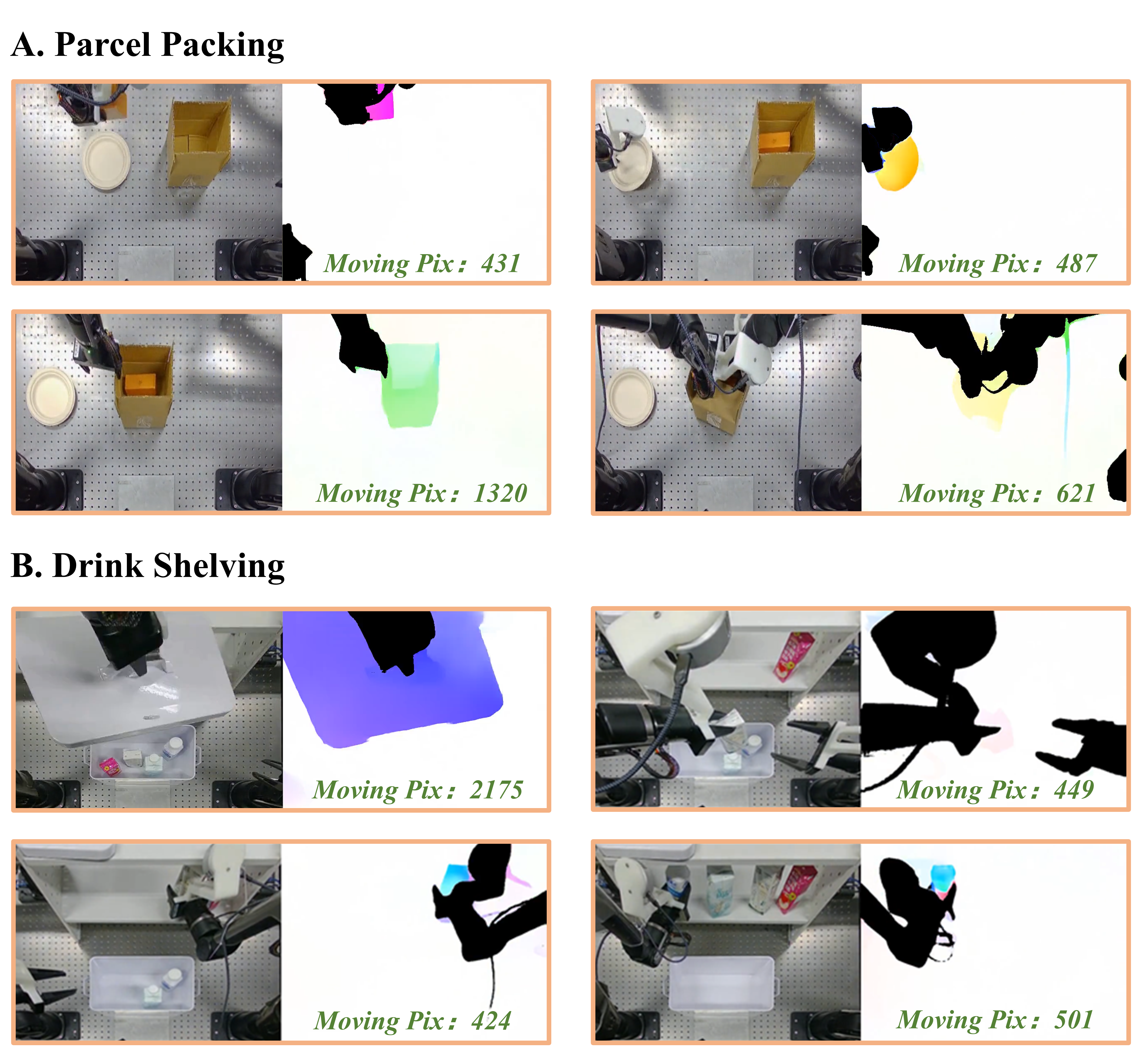}
    \caption{\textbf{Visualization of Interaction Signal Extraction.} We illustrate the process of quantifying interaction intensity. For each sample frame, the left panel shows the raw RGB observation from global camera view, while the right panel displays the dense optical flow with the robot body masked out (black region). The ``Moving Pix'' count represents the magnitude of environmental motion $\sum (1 - M_t) \cdot \mathbb{I}(||F_t||^2 > \delta_{\text{flow}})$, which serves as the threshold metric for determining the binary interaction signal $I_t$.}
    \label{fig:experiment_flow}
\end{figure}

First, we validate the effectiveness of our visual interaction extraction module in Figure~\ref{fig:experiment_flow}. By combining global RGB observations with dense optical flow and robot masking, our system accurately quantifies the intensity of environmental changes. Following Eq. 1, we determine the interaction status ($I_t$) by calculating the moving pixel count and applying a threshold $\Theta_{\text{pixel}} = 400$. The visualization highlights four representative frames selected from the interaction phase for both the Parcel Packing and Drink Shelving tasks. This method allows for the convenient and automatic extraction of interaction information. By reducing exploration noise during interaction phases to enhance stability and robustness, while employing higher variance exploration in non-interaction zones to foster diverse trajectories and generalization, this pipeline effectively balances exploration and exploitation in RL.

\begin{figure*}[t]
    \centering
    \includegraphics[width=2\columnwidth]{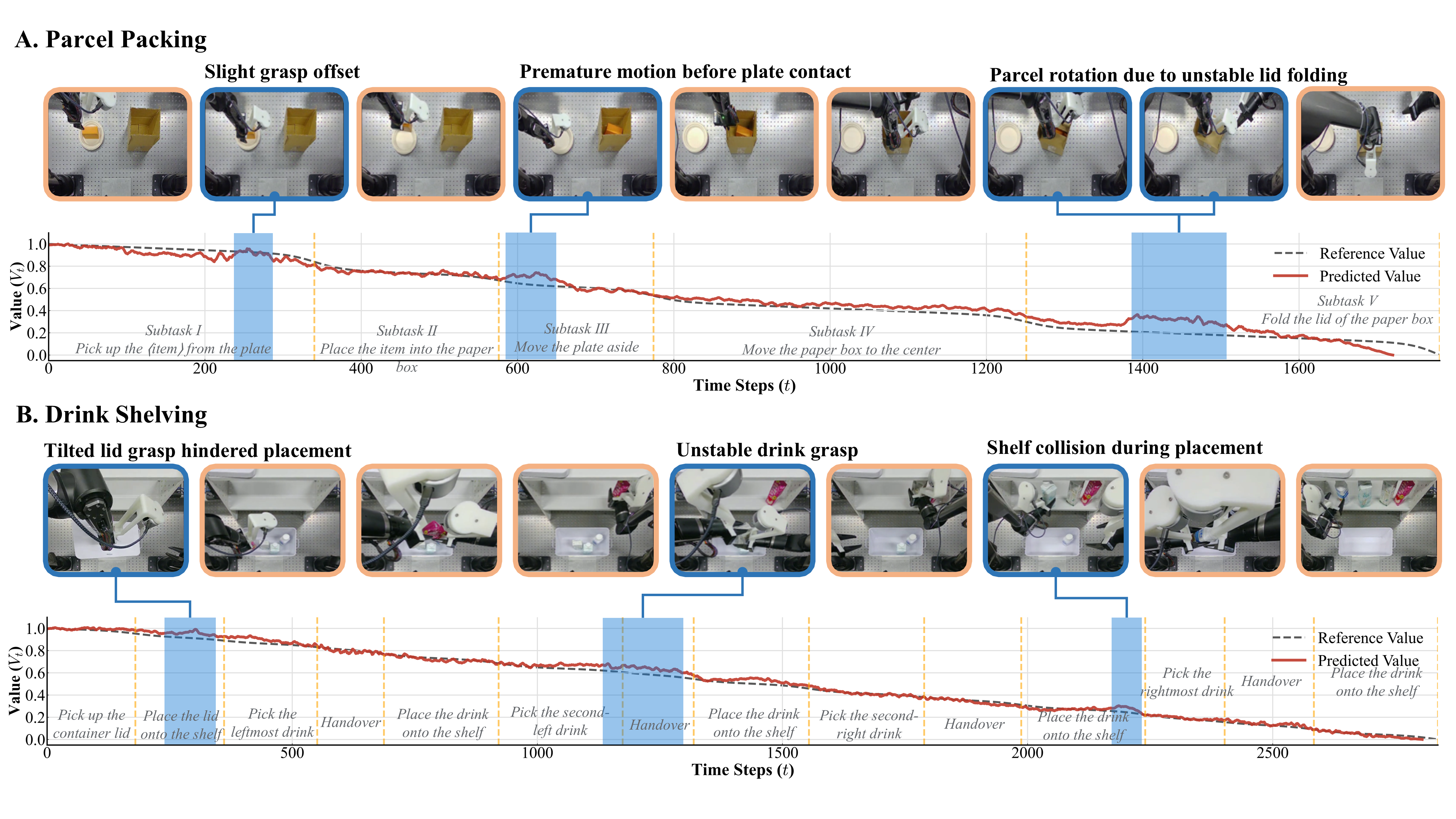}
    \caption{\textbf{Visualization of Value during one rollout in Long-Horizon Tasks.} We plot the critic-model-predicted state value (Return-to-Go) trajectories for \textbf{(A) Parcel Packing} and \textbf{(B) Drink Shelving} tasks. The dashed grey line represents the reference value function designed in Section IV-A, while the solid red line shows the predicted value during real-world execution. The images from left to right sequentially represent a successful task; the blue-highlighted images correspond to value peaks, indicating the policy took suboptimal actions. For each suboptimal case, we describe the potential reasons with text above the images. The corresponding subtask stages are annotated in grey italic text along the timeline.}
    \label{fig:experiment_value}
\end{figure*}

Furthermore, we analyze the critic's ability to evaluate policy performance by visualizing the predicted state value ($V_t$) trajectories in Figure~\ref{fig:experiment_value}. The red curve represents the critic's real-time prediction, while the dashed grey line indicates the reference value. In successful rollouts, the predicted value closely follows the reference, monotonically decreasing as subtasks are completed. Crucially, the critic exhibits high sensitivity to suboptimal actions. As highlighted by the blue shaded regions, the predicted value fluctuates or rises abnormally when the agent executes actions suboptimally, indicating potential risks rather than explicit failures. For instance, in the Parcel Packing task, a "slight grasp offset" or "premature motion" causes immediate value deviations. Similarly, in the Drink Shelving task, events like "unstable drink grasp" or "shelf collision" trigger distinct value peaks. These results demonstrate that our dense value function not only guides the policy towards the goal but also serves as an effective indicator of execution quality, providing critical dense feedback that significantly facilitates policy optimization during Reinforcement Learning.

\section{Conclusion}
\label{sec:conclusion}
In this work, we presented IG-RFT, a comprehensive reinforced fine-tuning system tailored for flow-based VLA models to master long-horizon complex manipulation tasks in the real world. To balance exploration and exploitation in RL training, we innovatively posit that interaction status contains strong prior guidance information for long-horizon contact-rich manipulation. Integrating this insight, we proposed the Interaction-Guided Advantage Weighted Regression (IG-AWR) algorithm, which dynamically modulates flow models at the noise level. Addressing the training difficulties caused by sparse rewards in RL and the fact that task-specific dense rewards require careful manual design and are hard to scale, we proposed a hybrid dense reward function. This function includes trajectory-level and subtask-level rewards, simultaneously achieving density and generalization.

Furthermore, our multi-stage training paradigm demonstrates a clear evolution of policy performance: establishing basic domain adaptation and value function estimation via SFT, optimizing to realize efficient policy learning via Offline RL, and achieving policy evolution while maintaining robust generalization through Human-in-the-Loop corrections. Real-world experiments across four challenging tasks validate that our method surpasses SFT and standard RL baselines by a significant margin, with approximately 40\% improvement. We also implemented comprehensive ablation experiments to validate our design choices.

As a powerful and effective policy improvement technique, RL possesses the potential to enable SFT models to break through the upper limit of their representation capability. However, existing research, including this work, tends to utilize RL as a fine-tuning method to bridge the domain gap and improve task success rates, rather than applying RL to large-scale training as seen in Large Language Models (LLMs) \cite{ouyang2022training, guo2025deepseek}. We attribute this to two main reasons: 1) Sampling and exploration using policy gradient-based online RL (like GRPO, PPO) \cite{shao2024deepseekmath, schulman2017proximal} in the high-dimensional action space of the real world are expensive, while simulation faces a significant sim-to-real gap, and Human-in-the-Loop guidance relies heavily on manual labor, making it difficult to scale up. 2) A general Reward Model that can be directly applied to fully on-policy, online, and new real-world domains does not yet exist; the lack of real-time feedback from such a model poses significant challenges for real-robot on-policy learning.

Future work will focus on two directions: applying RL to large-scale VLA post-training across multi-task and wide-ranging embodiments. Since human feedback currently remains an indispensable means, investigating how to efficiently provide feedback to enable robots to autonomously approach on-policy exploration holds great research promise. Additionally, training large reward models with world models—particularly through paradigms such as physics-aware video generation—represents a crucial pathway toward scalable RL for real-world robotics \cite{kim2026cosmos}.

\appendices

\section{Critic Network Architecture}
\label{sec:critic_architecture}
The critic module is designed based on an attention-based aggregation architecture inspired by Q-Former \cite{li2023blip}. To efficiently capture multimodal context, we initialize $N_q=8$ learnable query tokens, each with an embedding dimension of $D=256$. These queries interact with the frozen VLA features—extracted selectively from specific middle and late layers of the backbone (indices 11, 14, and 17 in our 18-layer model)—through a multi-head cross-attention mechanism with 8 heads. Note that we explicitly detach gradients from the VLA backbone to prevent interference with the policy representation. The aggregated features are then fed into three independent Multi-Layer Perceptron (MLP) heads. Specifically, each MLP head projects the $256$-dimensional feature to a hidden dimension of $128$ before producing a final scalar output. The heads are optimized over sampled batches $\mathcal{B}$ as follows:
\begin{enumerate}
    \item \textbf{Trajectory Value} $V_{\text{traj}}(s)$: Predicted by the task head to estimate long-term task success probability. It is optimized by minimizing the expected Mean Squared Error (MSE) against the ground truth return $y_{\text{traj}}$:
    \begin{equation}
        \mathcal{L}_{\text{traj}} = \mathbb{E}_{(s, y_{\text{traj}}) \sim \mathcal{D}} \left[ \left( V_{\text{traj}}(s) - y_{\text{traj}} \right)^2 \right]
    \end{equation}
    
    \item \textbf{Subtask Value} $V_{\text{sub}}(s)$: Predicted by the subtask head to track the progress of the current sub-goal. Similarly, it minimizes the expected MSE loss with respect to the subtask value label $y_{\text{sub}}$:
    \begin{equation}
        \mathcal{L}_{\text{sub}} = \mathbb{E}_{(s, y_{\text{sub}}) \sim \mathcal{D}} \left[ \left( V_{\text{sub}}(s) - y_{\text{sub}} \right)^2 \right]
    \end{equation}
    
    \item \textbf{Interaction Probability} $p_{\text{int}}(s)$: Predicted by the interaction head to identify contact-rich phases. It uses a Sigmoid activation and is supervised via the expected Binary Cross-Entropy (BCE) loss against the binary interaction label $y_{\text{int}}$:
    \begin{equation}
    \begin{split}
        \mathcal{L}_{\text{int}} = - \mathbb{E}_{(s, y_{\text{int}}) \sim \mathcal{D}} \Big[ & y_{\text{int}} \log p_{\text{int}}(s) \\
        & + (1 - y_{\text{int}}) \log (1 - p_{\text{int}}(s)) \Big]
    \end{split}
    \end{equation}
\end{enumerate}
The total critic loss is the summation of these three components: $\mathcal{L}_{\text{critic}} = \mathcal{L}_{\text{traj}} + \mathcal{L}_{\text{sub}} + \mathcal{L}_{\text{int}}$.

\section{Hyperparameters}
\label{sec:hyperparams}
We list the key hyperparameters used in our experiments in Table~\ref{tab:hyperparameters}. For the SFT stage, we use the standard flow matching objective. In the RL stages (Offline and Human-in-the-Loop), we utilize the proposed IG-AWR algorithm. The critic network is trained using a separate optimizer.

\begin{table}[h]
\caption{Hyperparameters for IG-RFT Training}
\label{tab:hyperparameters}
\centering
\renewcommand{\arraystretch}{1.1}
\resizebox{0.8\linewidth}{!}{
\begin{tabular}{l|c}
\toprule
\textbf{Parameter} & \textbf{Value} \\
\midrule
\multicolumn{2}{c}{\textit{General Optimization}} \\
Optimizer & AdamW \\
LR Schedule & CosineDecay \\
Learning Rate (Actor) & $5 \times 10^{-5}$ \\
Learning Rate (Critic) & $5 \times 10^{-6}$ \\
Batch Size & 64 \\
Gradient Clipping & 1.0 \\
EMA Decay & 0.999 \\
LR Warm-up Steps & 1,000 \\
Critic Warm-up Steps & 2,000 \\
Total Training Steps & 30,000 \\
\midrule
\multicolumn{2}{c}{\textit{IG-AWR Algorithm}} \\
AWR Temperature $\beta$ & 0.05 \\
Advantage Clipping & $[0, 10]$ \\ 
IG Flow Threshold $\delta_{\text{flow}}$ & 2.0 \\
IG Pixel Threshold $\Theta_{\text{pixel}}$ & 400 \\
Exploration $\sigma_{\text{base}}$ & 1.5 \\
Exploration $\sigma_{\text{min}}$ & 0.2 \\
Exploration Modulation $\alpha$ & 0.9 \\
Discount Factor $\gamma$ & 1.0 \\
Trajectory Reward Weight $\omega_{\text{traj}}$ & 0.4 \\
Subtask Reward Weight $\omega_{\text{sub}}$ & 0.6 \\
\midrule
\multicolumn{2}{c}{\textit{Training Schedule}} \\
SFT Warm-up Steps & 5,000 \\
Offline RL Training Steps & 25,000 \\ 
HitL RL Iterations & 4 \\
HitL RL Update Interval & 10 \\
Actor Update Frequency & 1.0 \\
Critic Update Frequency & 0.5 \\
\bottomrule
\end{tabular}
} 
\end{table}

\section{Task Descriptions}
\label{sec:tasks}
We provide detailed descriptions and strict success criteria for the four evaluation tasks to ensure experimental rigor.

\begin{itemize}
    \item \textbf{Parcel Packing:}
    \begin{itemize}
        \item \textit{Goal:} Place an item into a box and close the box.
        \item \textit{Success Criteria:} The object must be fully inside the box, and flaps must be folded flat (sealed).
        \item \textit{Subtasks Sequence:}
        \begin{enumerate}
            \item[1.] Pick up the $\langle \text{item} \rangle$ from the plate.
            \item[2.] Place the item into the paper box.
            \item[3.] Move the plate aside.
            \item[4.] Move the paper box to the center.
            \item[5.] Fold the lid of the paper box.
        \end{enumerate}
        \item \textit{Average Steps:} 1731.
    \end{itemize}
    \vspace{0.2cm}
    \item \textbf{Fruit Bagging:}
    \begin{itemize}
        \item \textit{Goal:} Put three specific fruits (ginger, bitter melon, banana) into a canvas bag and close it.
        \item \textit{Success Criteria:} All three items must be inside, and the lid must cover the opening.
        \item \textit{Subtasks Sequence:}
        \begin{enumerate}
        \item[1.] Open the lid of the canvas bag.
        \item[2.] \textbf{For each} $\langle \text{item} \rangle$ in $\{\text{ginger, bitter melon, banana}\}$:
            \begin{itemize}
            \item[$\triangleright$] Pick up the $\langle \text{item} \rangle$ from the tabletop.
            \item[$\triangleright$] Place the $\langle \text{item} \rangle$ into the canvas bag.
            \end{itemize}
        \item[3.] Move the canvas bag to the center.
        \item[4.] Close the lid of the canvas bag.
        \end{enumerate}
        \item \textit{Average Steps:} 2517.
    \end{itemize}
    \vspace{0.2cm}
    \item \textbf{Block Stacking:}
    \begin{itemize}
        \item \textit{Goal:} Stack 4 colored blocks in a specific order (Red $\to$ Yellow $\to$ Green $\to$ Blue).
        \item \textit{Success Criteria:} The stack must stand vertically for at least 3 seconds after release.
        \item \textit{Subtasks Sequence:}
        \begin{enumerate}
            \item[1.] Pick up the red block from the tabletop.
            \item[2.] Place the red block on the plate.
            \item[3.] \textbf{For each pairs} $\langle \text{source}, \text{target}         \rangle$ in sequence \\
              \hspace*{0.1em} $\{(\text{yellow}, \text{red}), (\text{green}, \text{yellow}), (\text{blue}, \text{green})\}$:
                \begin{itemize}
                \item[$\triangleright$] Pick up the $\langle \text{source} \rangle$ block from the tabletop.
                \item[$\triangleright$] Stack the $\langle \text{source} \rangle$ block on the $\langle \text{target} \rangle$ block.
            \end{itemize}
        \end{enumerate}
        \item \textit{Average Steps:} 1899.
    \end{itemize}
    \vspace{0.2cm}
    \item \textbf{Drink Shelving:}
    \begin{itemize}
        \item \textit{Goal:} Place the container lid on the top shelf, then retrieve 4 drinks from a bin and place them onto the middle shelf via dual-arm handover.
        \item \textit{Success Criteria:} The lid must be on the top shelf. All 4 drinks must be placed upright on the middle shelf layer without falling.
        \item \textit{Subtasks Sequence:}
        \begin{enumerate}
            \item[1.] Pick up the container lid from the bin.
            \item[2.] Place the lid onto the top shelf.
            \item[3.] \textbf{For each} $\langle \text{loc} \rangle$ in sequence \\
                \hspace*{0.1em} $\{\text{leftmost}, \text{second-left}, \text{second-right}, \text{rightmost}\}$:
               \begin{itemize}
                 \item[$\triangleright$] Pick up the $\langle \text{loc} \rangle$ drink from the bin using a top-down grasp.
                 \item[$\triangleright$] Hand over the drink to the other arm.
                 \item[$\triangleright$] Place the drink onto the middle shelf via side-insertion.
               \end{itemize}
        \end{enumerate}
        \item \textit{Average Steps:} 2931.
     \end{itemize}
\end{itemize}

\bibliographystyle{IEEEtran}
\bibliography{reference} 

\end{document}